\documentclass[letterpaper, 10 pt, conference]{ieeeconf}  

\IEEEoverridecommandlockouts                              

\usepackage[pdftex]{graphicx}
\usepackage{amsmath}
\usepackage{amssymb}  
\usepackage{subfigure}
\usepackage{multirow, makecell}
\usepackage{array,booktabs}
\usepackage{balance}  
\usepackage{adjustbox}
\usepackage{tikz} 
\usepackage{xcolor,colortbl}

\makeatletter
\let\NAT@parse\undefined
\makeatother
\usepackage[numbers]{natbib}

 \usepackage{hyperref}
 \hypersetup{
     colorlinks=true,
     linkcolor=blue,
     filecolor=magenta,
     urlcolor=blue,
     citecolor=black
 }

\providecommand{\figref}[1]{Fig.~\ref{#1}}
\providecommand{\tabref}[1]{Table~\ref{#1}}

\providecommand{\secref}[1]{\textsection\ref{#1}}

\long\def\symbolfootnote[#1]#2{\begingroup%
\def\thefootnote{\fnsymbol{footnote}}\footnote[#1]{#2}\endgroup}

\usepackage{acronym}
\acrodef{SLAM}{simultaneous localization and mapping}

\usepackage{soul,color}
\usepackage{lipsum}

\usepackage{xcolor}
\definecolor{darkgreen}{HTML}{32CD32}
\definecolor{Gray}{gray}{0.9}

\newcommand{\B}[1]{{\textbf{#1}}}
\newcommand{\BK}[1]{{\textbf{\textcolor{black}{#1}}}}

\newcommand{\id}[1]{{#1}}

\begin{document}

\title{\LARGE \bf
Fieldscale: Locality-Aware Field-based Adaptive Rescaling \\for Thermal Infrared Image
}

\author{Hyeonjae Gil$^{1}$, Myung-Hwan Jeon$^{1}$, and Ayoung Kim$^{1*}$%
\thanks{$^{1}$H. Gil, M. Jeon and A. Kim are with the Department of Mechanical Engineering, SNU, Seoul, S. Korea {\tt\footnotesize [h.gil, myunghwan.jeon, ayoungk]@snu.ac.kr}}%
}

\maketitle

\begin{abstract}
Thermal infrared (TIR) cameras are emerging as promising sensors in safety-related fields due to their robustness against external illumination.
However, RAW TIR image has 14 bits of pixel depth and needs to be rescaled into 8 bits for general applications.
Previous works utilize a global 1D look-up table to compute pixel-wise gain solely based on its intensity, which degrades image quality by failing to consider the local nature of the heat. 
We propose Fieldscale, a rescaling based on locality-aware 2D fields where both the intensity value and spatial context of each pixel within an image are embedded.
It can adaptively determine the pixel gain for each region and produce spatially consistent 8-bit rescaled images with minimal information loss and high visibility. 
Consistent performance improvement on image quality assessment and two other downstream tasks support the effectiveness and usability of Fieldscale.
All the codes are publicly opened to facilitate research advancements in this field. \hyperlink{https://github.com/hyeonjaegil/fieldscale}{https://github.com/hyeonjaegil/fieldscale}
\end{abstract}

\section{Introduction}
\label{sec:intro}
Despite the widespread use of RGB cameras in the perception of autonomous vehicles and robotics \cite{geiger2013vision, jeon2022ambiguity}, their sensitivity to changes in lighting conditions can lead to safety concerns.
Some lines of work attempted to solve it through low light enhancement \cite{li2021learning} and adaptive exposure control \cite{kim2020proactive}, but using complementary novel sensors can be a fundamental solution to effectively support or improve these limitations.
In this context, thermal infrared (TIR) cameras can serve as a suitable alternative, capable of measuring long-wave infrared (LWIR) radiation without being affected by external factors, including light sources or climate changes.
Compared to other ranging sensors, TIR imaging has the advantage of facilitating the transfer of research in object detection \cite{kim2023transpose}, semantic segmentation \cite{sun2019rtfnet}, or place recognition \cite{saputra2021graph} within the RGB domain.

\begin{figure}[!t]
    \centering
   \vspace{-1mm}
    \subfigure[Different characteristics between RGB and TIR imaging]{
		\includegraphics[width=0.9\columnwidth]{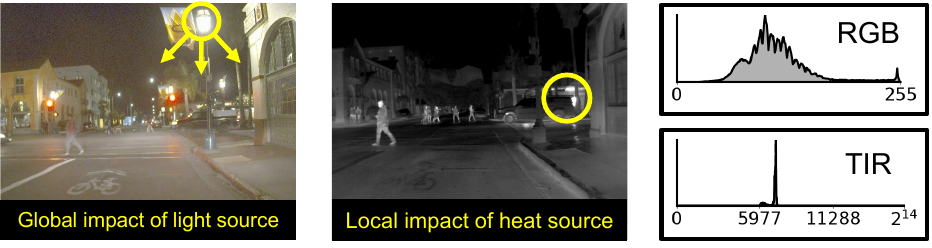}\label{fig:rgb_comparison_a}}
    \subfigure[Comparison with previous 1D LUT-based methods]{
		\includegraphics[width=0.9\columnwidth]{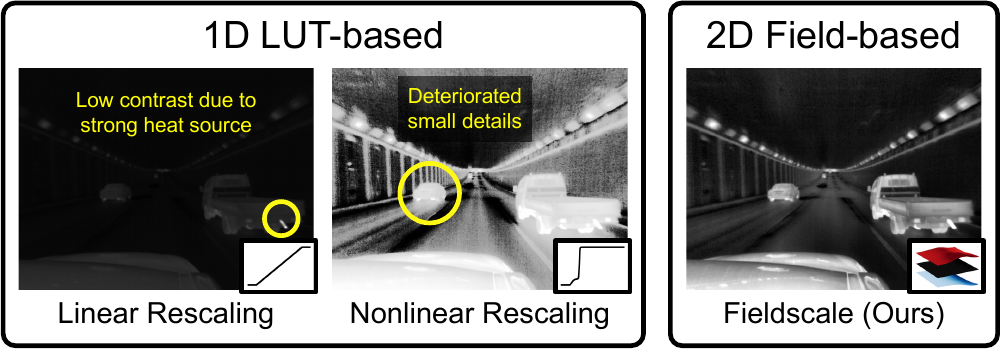}\label{fig:rgb_comparison_b}}
    \vspace{-1mm}
    \caption{Due to the locality of the heat source, rescaling the thermal image requires a spatially adaptive strategy. 
    In contrast to previous 1D LUT-based methods, Fieldscale can successfully handle local heat sources and enhance both global and local details.}
    \label{fig:rgb_comparison}
    \vspace{-5mm}
\end{figure}

Unfortunately, RAW TIR images are characterized by a pixel depth of 14 bits, which restricts their immediate applicability. 
While thermal odometry \cite{chen2023thermal, wang2023edge} may utilize RAW intensity for radiometric feature tracking, applications reliant on image appearance have been proven to exhibit improved robustness when using 8-bit TIR imagery \cite{saputra2021graph}. 
As shown in \figref{fig:rgb_comparison_a}, TIR images typically utilize a significantly narrower range than their full 14-bit spectrum; therefore, rescaling the occupied intensity range to an 8-bit spectrum is crucial for further TIR image-based applications.
As the rescaling process begins every workflow, minimizing information loss is key to enhancing the performance of subsequent applications.

Similar to the global tone mapping in RGB high dynamic range (HDR) imaging \cite{reinhard2010high, 15realtime}, previous TIR rescaling utilizes global 1D look-up tables (LUTs) that associate the RAW intensity of each pixel with a specific gain. These methods are generally divided into linear and nonlinear. In linear rescaling, a linear LUT applies uniform gain across all pixels, preserving original radiometry. This approach, however, decreases gain and global contrast in the presence of locally heated objects, as shown in \figref{fig:rgb_comparison_b}. Conversely, nonlinear rescaling adjusts gain based on pixel distribution using an image histogram, effectively enhancing contrast but potentially reducing detail in less populated pixel areas, particularly near small heated objects. 
Some hardware manufacturers offer advanced nonlinear rescaling software, such as automatic gain control (AGC) in FLIR cameras. While this improves image quality, the dependence on specific hardware limits reproducibility and scalability.

We argue that existing methods have overlooked the unique local properties of TIR imaging, which differ from those of RGB imaging. 
RGB cameras capture light reflected from external sources, typically resulting in a scene with relatively uniform brightness. 
Such uniformity allows for the use of global 1D LUTs, such as gamma curves, to compress HDR images into a low dynamic range (LDR) format. 
Thermal radiation, on the other hand, depends on the inherent properties of objects, with local heat sources impacting only their immediate vicinity. 
It leads to extreme spatial temperature variance and raises questions about the effectiveness of a global 1D LUT, which can obscure essential thermal details. 
Consequently, TIR imaging needs a more comprehensive rescaling strategy encompassing both the pixel values and the spatial context within the image. 

In this paper, we propose Fieldscale, a 2D field-based rescaling tailored to TIR images with high spatial temperature variance. 
Fieldscale utilizes two scalar fields, each matching the size of the image and incorporating the local information of each area.
At its core, we formulate the field construction as a graph construction problem, with information on each image area serving as nodes.
We also leverage a message passing scheme from a Graph Neural Network (GNN) to generate spatially consistent and natural images.
Consequently, Fieldscale can adaptively determine the gain of each pixel appropriate to the spatial distribution of temperature.
It can yield images with maximized contrast, rich details, and minimal information loss regardless of the temperature or size of an object.
To this end, Fieldscale can enhance the performance of subsequent tasks that rely on image quality or detailed local information.
Our contributions can be summarized as follows:

\begin{itemize}
    \item We introduce Fieldscale, a locality-aware 14-bit to 8-bit TIR rescaling methodology, 
    in which each pixel adapts its gain based on the spatial context. 
    This method maintains local radiometry, fine detail, global context, and spatial consistency.
    Its single-shot and real-time operation capability makes it versatile for applications ranging from visualization to safety-related tasks.
    \item The Fieldscale features a streamlined field construction process with minimal and insensitive parameters. 
    Its modular design allows easy customization of each component to meet specific needs. 
    Moreover, it facilitates the incorporation of additional image enhancements for further quality improvement, 
    making Fieldscale a scalable alternative to previous 1D LUT-based methods.
    \item Effectiveness of Fieldscale is proven through superior results in image quality assessment metrics. 
    It has also been thoroughly validated in various downstream applications, such as visual place recognition and object detection. 
    Every codebase is open-sourced to accelerate future research on TIR imaging within robotics.
\end{itemize}

\section{Related Works}
\label{sec:relatedwork}

\subsection{Thermal Image Rescaling}
Linear rescaling, also known as fixed interval rescaling or clipping, selects two fixed endpoints to clamp and rescale the original image. They can be either set from specific temperatures \cite{shin2019sparse}, statistical percentile values \cite{23shin}, or arbitrary values \cite{khattak2020keyframe}, but the important information from the objects with extreme temperatures can easily be washed out.
Nonlinear rescaling mainly consists of histogram equalization (HE) variants.
\citeauthor{9804833} \cite{9804833} modified HE with 30 bins and applied contrast limited adaptive histogram equalization (CLAHE), but the detail from a small portion of intensity can be lost.
On top of logarithmic compression, multiscale retinex (MSR) \cite{97msr}-based approaches have also been suggested.
\citeauthor{15cgf} \cite{15cgf} compressed the intensity into the log domain and subtracted the images filtered with multiscale conditional Gaussian kernels. To save the long processing time of MSR-based schemes, \citeauthor{23joint} \cite{23joint} emulated MSR with additional image enhancement by trained Deep Neural Network (DNN) in a supervised manner, but it requires 8-bit rescaled images for ground-truth.

\subsection{Thermal Image Enhancement}
Various efforts have addressed the issues specific to TIR imaging. Thermal inertia causes blurring in microbolometer-based TIR cameras, where \citeauthor{ramanagopal2020pixel} \cite{ramanagopal2020pixel} proposed pixel-wise deblurring to reverse this effect by modeling it as an underdetermined system of equations and using temporal sparsity constraints. Non-uniform spatial bias generated by the noise in the microbolometer is another intrinsic problem. \citeauthor{jiang2022thermal} \cite{jiang2022thermal} introduced SVD-based components to remove strip noise. Spatiotemporal calibration of pixel gain \cite{das2021online} improved the feature tracking \cite{polizzi2022data}, but it requires accurate pixel correspondences from the pre-calibrated images.

\subsection{Robotics Application of Thermal Images}
Thermal imaging has been researched across various fields as a standalone or multi-modal approach, addressing environments that conventional RGB images alone may struggle. \citeauthor{munir2021sstn} \cite{munir2021sstn} proposed a self-supervised multi-modal domain adaptation framework with contrastive learning for thermal object detection in autonomous driving.
With the aim of wilderness search and rescue, \citeauthor{broyles2022wisard} \cite{broyles2022wisard} introduced a large-scale dataset labeled visual and thermal images. \citeauthor{polizzi2022data} \cite{polizzi2022data} proposed a decentralized, collaborative thermal-inertial odometry system for multiple UAVs that enables robust state estimation and navigation in dark extraterrestrial environments such as Mars. Regardless of the type or field of application, there is a common need for 8-bit TIR images with sufficient information and details.

\section{Methods}
\label{sec:methods}

\begin{figure*}[!t]
    \centering
   \vspace{-1mm}
    \subfigure[Fieldscale overview]{
        \includegraphics[width=0.95\linewidth]{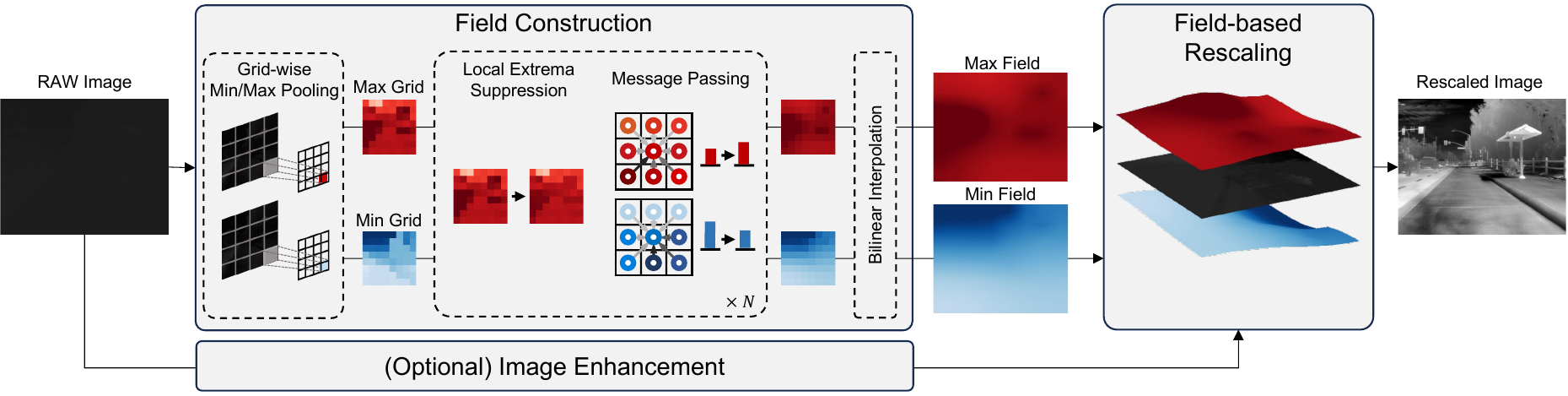}
        \label{fig:overview}
        } 
    \\
    \subfigure[Scene with large sky (cold area)]{        
        \includegraphics[width=0.3\linewidth]{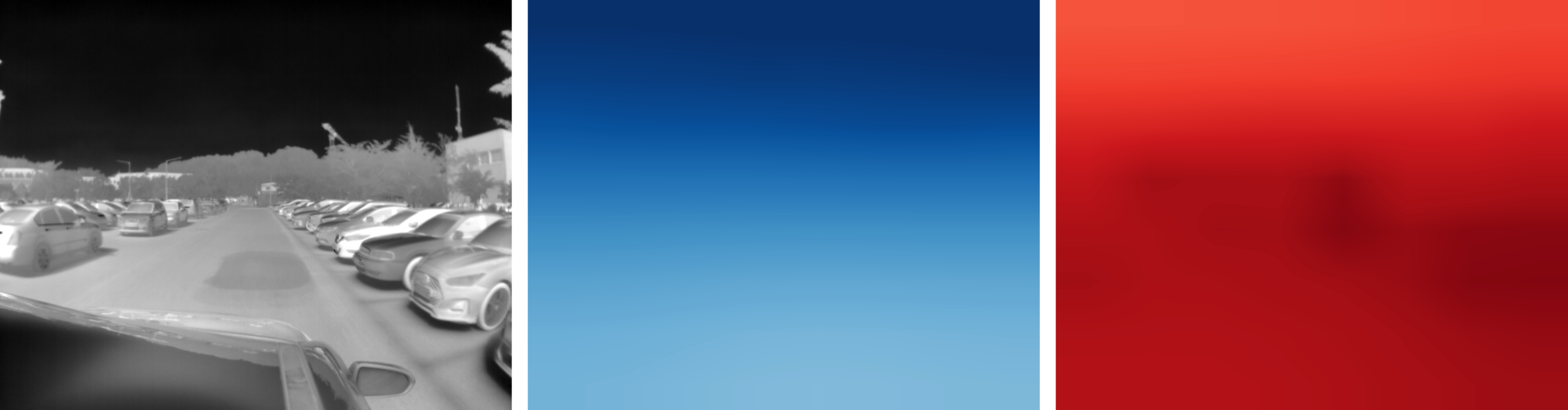}
        \label{fig:image_with_fields_scene1}
        }
    \subfigure[Scene with local hot objects]{        
        \includegraphics[width=0.3\linewidth]{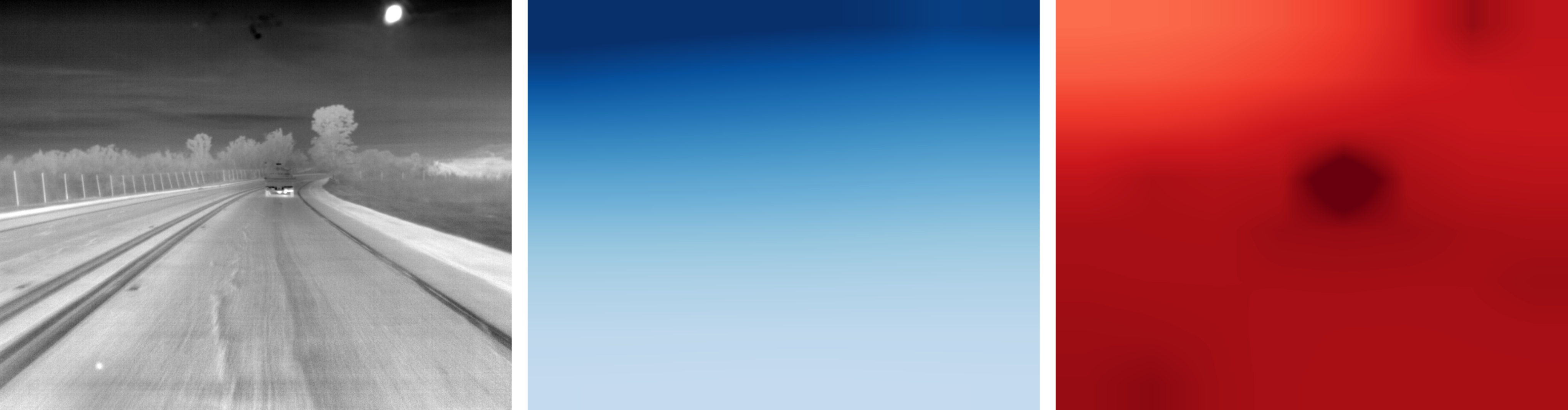}
        \label{fig:image_with_fields_scene2}
        }
    \subfigure[Scene with small local details]{        
        \includegraphics[width=0.3\linewidth]{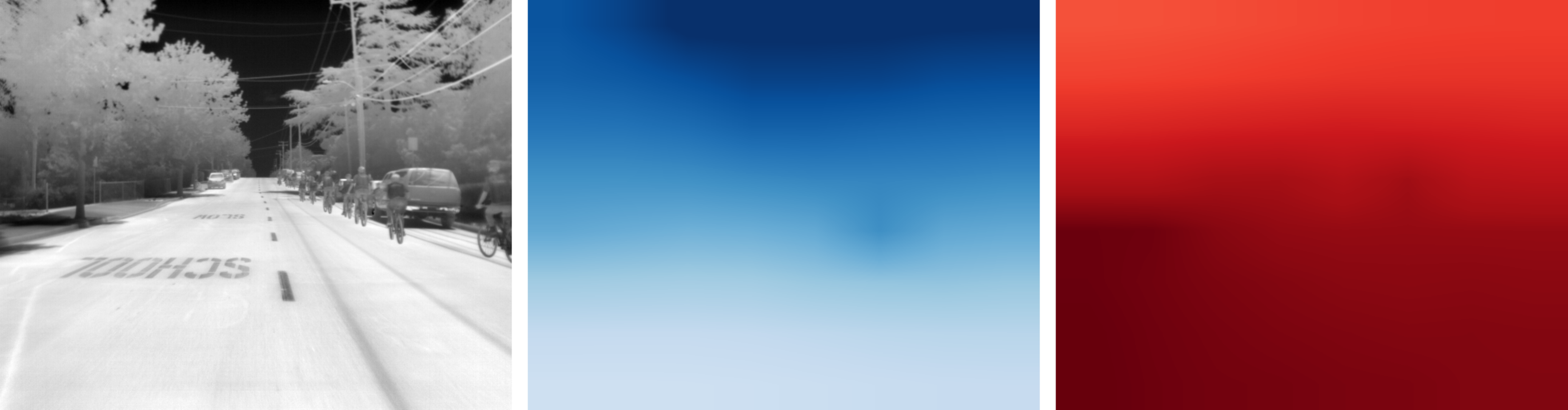}
        \label{fig:image_with_fields_scene3}
        }
    \vspace{-1mm}
    \caption{Overview of Fieldscale and examples of rescaled images with corresponding fields. By passing the RAW image into three sequential modules (Grid-wise Min/Max Pooling, Local Extrema Suppression, and Message Passing), min/max fields with the same size of the image are constructed from the local information of each area. A darker pixel indicates the extreme values: a lower value for the min field and a higher value for the max field. 
    After applying optional image enhancement in parallel, the RAW image is rescaled with two fields.
    As visualized in three examples, cold regions bring down the corresponding area of the min field, and hot regions raise that of the max field. This locally adaptive field construction can successfully rescale images with high spatial temperature variance without loss of original details.}
    \label{fig:overall_diagram}
    \vspace{-5mm}
\end{figure*}

This section introduces the motivation, overview, and components of our field-based rescaling approach.
We also report the scalability of our system with the possibility of integrating other thermal image enhancement modules. 

\subsection{From Transfer Function to Scalar Fields}
As mentioned in \secref{sec:intro}, it is possible to generalize previous rescaling into designing a 1D LUT, or namely \textit{transfer function}.
Transfer function $T: \mathbb{N} \rightarrow \mathbb{N}$ maps the RAW 14 bits intensity of input TIR image into 8 bits output image $O(x,y) = T(I(x,y))$. 
We first focus on linear rescaling, which requires only two scalar values, global minimum and maximum values ($\text{min}_{global}$ and $\text{max}_{global}$), to construct a linear transfer function.
It can be expressed as:
\begin{equation}
O(x,y) = CC_{0,255}(255 \times \frac{I(x,y) - \text{min}_{global}}{\text{max}_{global} - \text{min}_{global}})
\end{equation}
where  $CC_{0,255}(\cdot)$, namely \textit{clamp-and-convert}, is an operator that quantizes input into an integer, clamps each pixel value to 0 and 255, and then converts the data type to \texttt{uint8}. 

Within linear rescaling, all pixels share the same scalar values ($\text{min}_{global}$ and $\text{max}_{global}$).
This is analogous to using two scalar fields $\phi_{min}, \phi_{max} \in \mathbb{N}^{2}$ whose sizes are identical to the image $I(x,y)$ with values filled with $\text{min}_{global}$ and $\text{max}_{global}$, respectively.
Using scalar fields, the same linear rescaling with scalar values can be re-expressed as follows:
\begin{equation}
O(x,y) = CC_{0,255}(255 \times \frac{I(x,y) - \phi_{min}(x,y)}{\phi_{max}(x,y) - \phi_{min}(x,y)})
\end{equation}
That is, rescaling with a single linear transfer function is equivalent to rescaling with two scalar fields.
This representation allows us to utilize not only pixel intensity values but also its spatial information, i.e., locality. 
Analogous to previous rescaling where global information such as histogram is utilized to design the best 1D transfer function, 
we aim to design two 2D scalar fields with local information.

\subsection{Locality-aware Field Construction}
\figref{fig:overview} illustrates the field construction of Fieldscale, where three sequential modules followed by bilinear interpolation produce two scalar fields.

\subsubsection{Grid-wise Min/Max Pooling}
We first divide an image into uniform $N \times N$ areas and then pool the minimum and maximum pixel values within each patch to create $N \times N$ sized \textit{min grid} and \textit{max grid}. 
This offers various advantages: 
\begin{itemize}
   \item constraining the influence of objects with extreme temperature on the patch
   \item disentanglement of processing cold and hot areas
   \item faster subsequent processing regardless of image size
\end{itemize}
Regarding the grid size, a smaller size would make it hard to encode sufficient local information, while a bigger size would increase the time consumption. 
We empirically find that $8\times8$ grid size is the sweet spot for outdoor scenes and $4\times4$ for indoor scenes with a much narrower intensity range.

\subsubsection{Local Extrema Suppression (LES)}
Min grid and max grid can hold the local minimum and maximum information of each region, but their values could significantly differ from their surroundings if there are extremely hot or cold objects in the local area. 
This necessitates additional handling to hinder it from deteriorating the overall image appearance.
As a remedy, we first view the entire grid as a graph $G=(V, E)$, with each element of a grid as a node $V$, and $E$ as the set of edges connecting each node to its 8-neighboring elements. 
For each node $v\in V$, we also define a neighborhood function $\mathcal{N}(v,d)$, which returns the set of nodes within a distance $d$, and its average value $A(v,d)$.
We apply local extrema suppression (LES) to every node in the graph as follows:
\begin{equation}
LES(v,d) = 
\begin{cases} 
  A(v,d) - T_{\text{LES}}, & \text{if } v < A(v,d) - T_{\text{LES}} \\
  A(v,d) + T_{\text{LES}}, & \text{if } A(v,d) + T_{\text{LES}} < v \\
  v, & \text{otherwise}
\end{cases} 
\end{equation}
where $T_{\text{LES}}$ is the LES threshold value.
We fix $d=2$ for $8\times8$ grid size.
As seen in the dark pixels of \figref{fig:image_with_fields_scene2}, LES can adequately clamp the value of the max grid where the objects with extreme temperatures are located.
In our datasets, the min grid rarely experiences locally extreme values, so LES is only applied to the max field.

\subsubsection{Message Passing (MP)}
Inspired by message passing (MP) in GNN in which adjacent nodes exchange embedding, we propose an MP scheme for the min/max grids. 
It does not exchange embedding \textit{vectors} or use neural networks; each pixel exchanges its min/max \textit{scalars}. 
Similar to GNN, aggregation ($\operatorname{Agg}$) and update ($\operatorname{Update}$) functions are defined to update the value of each node:
\begin{equation}
v^{t+1}
=\operatorname{Update}\left(
v^t, \operatorname{Agg}\left(\mathcal{N}(v,1)\right)
\right)
\end{equation}
$t$ denotes the current iteration of MP (total $N$ iterations).
We use an averaging operator with distance $d=1$ for $\operatorname{Agg}$.
For $\operatorname{Update}$, we use the $\text{min}(x,y)$ operator for the min grid to moderately decrease the value and the $\text{max}(x,y)$ operator for the max grid to increase the value for every iteration.
Although these operators are fixed in the rest of the paper, they can easily be customized for specific downstream tasks.

\subsubsection{Bilinear Interpolation}
After passing through the three components, the resulting grid is upscaled to the same size as the image using bilinear interpolation. Similar to \cite{9804833}, we subsequently apply gamma correction with $\gamma=1.5$ and CLAHE after rescaling to maximize the local contrast.

\subsection{Parallel processing with Image Enhancement Modules}

Fieldscale not only enriches local details but also the overall appearance and contrast. However, depending on the purpose, additional image enhancement modules such as sharpening, deblurring \cite{ramanagopal2020pixel}, or non-uniform bias correction \cite{das2021online} can be considered. Unlike traditional rescaling which applies enhancement to 8-bit rescaled images sequentially, Fieldscale allows them in parallel, offering significant benefits in preventing information loss due to 8-bit quantization.
As shown in \figref{fig:overview}, the RAW image is simultaneously sent to the field construction and enhancement module, and the processed raw image is rescaled with the fields. 
Fieldscale can also be integrated into the online spatial bias calibration pipeline \cite{das2021online} by offering accurate pixel correspondences from high-contrast images and receiving the bias correction results.
Parallel enhancements were not required and thus not applied in our experiments, but they can be utilized in future research whenever needed.

\section{Experimental Results}
\label{sec:experiment}

\begin{figure}[!b]
\centering
\includegraphics[width=0.9\columnwidth]{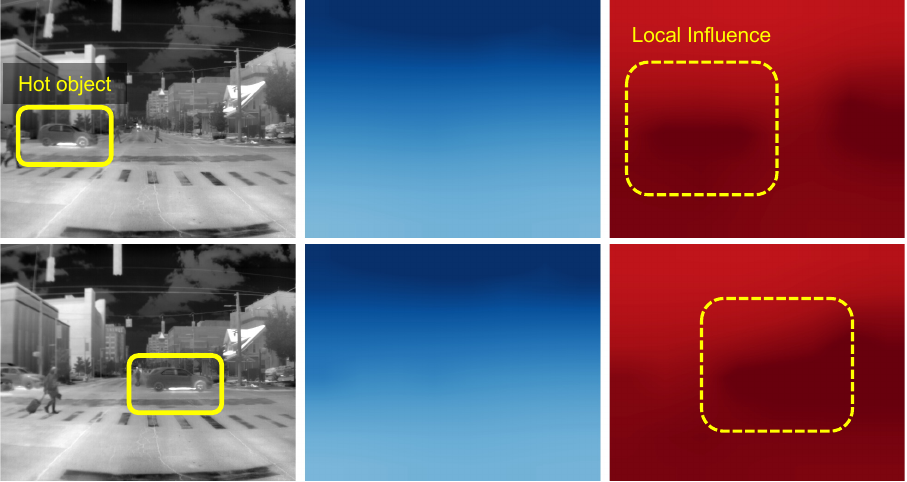}
\caption{
A hot object only influences the corresponding area of the max field while leaving the min field unaffected.
The local neighborhood information solely determines the pixel gain during the rescaling process.
}
\label{fig:locality}
\end{figure}

To evaluate our rescaling method, quantitative and qualitative image quality assessments are performed, followed by an impact analysis of two tunable parameters. 
Image-wise evaluation results are further supported by the results of two applications, visual place recognition and object detection.

\subsection{Qualitative Evaluation}
\subsubsection{Locality characteristics of Fieldscale}
We qualitatively evaluate the spatially adaptive characteristic of Fieldscale in \figref{fig:locality}, in which moving hot objects are captured from the static vehicle waiting for the traffic.
No change in the intensities of the min field indicates the disentanglement of processing hot and cold areas.
Hot objects only raise the corresponding area of the max field, but the remaining parts are not updated.
In other words, each pixel value of the rescaled image is solely determined by that of its neighborhood.
Owing to this locality, high spatial temperature variance can be successfully handled to produce rescaled images with minimum detail loss in every region.

\subsection{Quantitative Evaluation}
\subsubsection{IQA metrics}
Although there are several image quality assessment (IQA) metrics for image-level evaluation in the RGB domain, many of them are either \textit{full-reference} approaches \cite{wang2004image} requiring 8 bits images to compare with or neural network-based methods \cite{ke2021musiq} trained on RGB images.
So we leverage \textit{no-reference} IQA metrics (gradient and entropy) from low-light vision evaluation \cite{kim2020proactive}, and TIR-specific metrics from \cite{teutsch2020evaluation} which compares 14 bits RAW and 8-bit rescaled images comprehensively.
In specific, we measure global loss of contrast ($C_{\text{global}}$), local loss of contrast ($C_{\text{local}}$), and tone mapping quality index (TMQI) for the rescaling. TMQI quantifies multi-scale structural fidelity and statistical naturalness measures.

\subsubsection{Datasets and Baselines}
We perform IQA on three different sequences; validation images of FLIR ADAS dataset \cite{fliradas} (\texttt{FLIR-Val}), R0-FA0 daytime sequence of NSAVP dataset \cite{24nsavp} (\texttt{NSAVP-Day}), and campus night sequence of ViViD++ dataset \cite{lee2022vivid++} (\texttt{ViViD-Night}).
All three sequences were captured at different times, places, and camera models (FLIR Tau2, Boson, and A65, respectively), providing generalizability of our IQA results.
All frames in \texttt{FLIR-Val} are used, and both \texttt{NSAVP-Day} and \texttt{ViViD-Night} are subsampled so that every frame is 10m apart from others.
Baseline rescaling methods can be categorized as below:
\begin{itemize}
 \setlength\itemsep{0.8mm}
\item \underline{FLIR AGC}: AGC result of FLIR software
\item \underline{Minmax}: Linear rescaling with global min/max values
\item \underline{Clipping-I}: Clipping with 1\% and 99\% percentile values of the image
\item \underline{Clipping-V} \cite{23shin}: Clipping with average 1\% and 99\% percentile values of all images in a video
\item \underline{Shin-HE} \cite{9804833}: Modified version of histogram equalization with 30 bin size followed by CLAHE
\item \underline{MSR} \cite{97msr}: Multi-scale Retinex with $\sigma=[15,80,250]$
\item \underline{CGF} \cite{15cgf}: Conditional Gaussian Filter followed by clipping with 3$\sigma$ values of the image
\item \underline{JointTMO} \cite{23joint}: DNN-based jointly rescaling and denoising, which is trained with 8-bit reference optimized MSR images
\end{itemize}
CGF is faithfully re-implemented, and JointTMO is evaluated with the trained weights provided by the authors.
FLIR-AGC is only available in \texttt{FLIR-Val} and Clipping-I is not included in non-video sequences. We use $N$=7 and $T_{\text{LES}}$=100 for Fieldscale.

\subsubsection{Discussion}

\begin{table}[!t]
\centering
\caption{Image quality assessment results of each rescaling method
(\B{bold}: the best, \ul{underline}: the second best)}
\begin{adjustbox}{width=1\linewidth}
\begin{tabular}{@{\extracolsep{2pt}}llccccc}
\toprule[1.2pt]
\multirow{2}{*}{}  & \multirow{2}{*}{Rescaling} &  \multicolumn{2}{c}{Image-level}  &   \multicolumn{3}{c}{Rescaling-level} \\    \cline{3-4} \cline{5-7}
                            &                           & Gradient($\uparrow$) & Entropy($\uparrow$) & $C_{\text{global}}$($\downarrow$) & $C_{\text{local}}$($\downarrow$) & TMQI($\uparrow$)\\
\midrule[1.2pt]
\multirow{8}{*}{\rotatebox[origin=c]{90}{\texttt{FLIR-Val} \cite{fliradas}}}
& FLIR AGC                  & 0.0579        & \B{0.8729}    & -0.1338       & \ul{-0.0446}  & \ul{0.9329} \\
& Minmax                    & 0.0111        & 0.4137        & -0.0341       & -0.0210       & 0.7551 \\
& Clipping-I                & 0.0189        & 0.5520        & -0.0449       & -0.0128       & 0.7958 \\
& MSR \cite{97msr}          & 0.0141        & 0.4809        & -0.0394       & -0.0238       & 0.7562 \\
& CGF \cite{15cgf}          & 0.0247        & 0.6806        & -0.0653       & -0.0218       & 0.8475 \\
& JointTMO \cite{23joint}   & \ul{0.0671}   & 0.7721        & \B{-0.1749}   & -0.0332       & 0.9041 \\
& Shin-HE \cite{9804833}    & 0.0430        & 0.7767        & -0.0864       & -0.0332       & 0.9322 \\
& \B{Fieldscale (Ours)}     & \B{0.0679}    & \ul{0.8586}   & \ul{-0.1558}  & \B{-0.0513}   & \B{0.9444} \\
\midrule
\multirow{8}{*}{\rotatebox[origin=c]{90}{\texttt{NSAVP-Day} \cite{24nsavp}}}
& Minmax                    & 0.0098        & 0.4304        & -0.0365       & -0.0293       & 0.7490 \\
& Clipping-I                & 0.0179        & 0.5959        & -0.0446       & -0.0162       & 0.8085 \\
& Clipping-V \cite{23shin}  & 0.0174        & 0.5642        & -0.0447       & -0.0156       & 0.7998 \\
& MSR \cite{97msr}          & 0.0120        & 0.4937        & -0.0412       & -0.0343       & 0.7465 \\
& CGF \cite{15cgf}          & 0.0227        & 0.5631        & -0.0479       & -0.0237       & 0.8458 \\
& JointTMO \cite{23joint}   & \B{0.0694}    & \ul{0.7799}   & \B{-0.1816}   & -0.0355       & 0.8992 \\
& Shin-HE \cite{9804833}    & 0.0344        & 0.4921        & -0.0805       & \ul{-0.0355}  & \ul{0.9176} \\
& \B{Fieldscale (Ours)}     & \ul{0.0583}   & \B{0.8343}    & \ul{-0.1508}  & \B{-0.0490}   & \B{0.9481} \\
\midrule
\multirow{8}{*}{\rotatebox[origin=c]{90}{\texttt{ViViD-Night} \cite{lee2022vivid++}}}
& Minmax                    & 0.0120        & 0.5273        & -0.0248       & -0.0107       & 0.7892 \\
& Clipping-I                & 0.0145        & 0.5676        & -0.0339       & -0.0085       & 0.7896 \\
& Clipping-V \cite{23shin}  & 0.0141        & 0.5375        & -0.0304       & -0.0076       & 0.7807 \\
& MSR \cite{97msr}          & 0.0126        & 0.5549        & -0.0233       & -0.0121       & 0.7697 \\
& CGF \cite{15cgf}          & 0.0162        & 0.6181        & -0.0347       & -0.0176       & 0.8018 \\
& JointTMO \cite{23joint}   & \ul{0.0468}   & \ul{0.7637}   & \ul{-0.1112}  & \ul{-0.0316}  & 0.9099 \\
& Shin-HE \cite{9804833}    & 0.0382        & 0.5710        & -0.0711       & -0.0255       & \ul{0.9184} \\
& \B{Fieldscale (Ours)}     & \B{0.0533}    & \B{0.8402}    & \B{-0.1249}   & \B{-0.0515}   & \B{0.9459} \\
\bottomrule[1.2pt]
\end{tabular}
\end{adjustbox}
\label{tab:iqa}
\vspace{-2mm}
\end{table}

\begin{figure}[!t]
    \centering
    \subfigure[\texttt{FLIR-Val} \cite{fliradas}]{\includegraphics[width=1\linewidth]{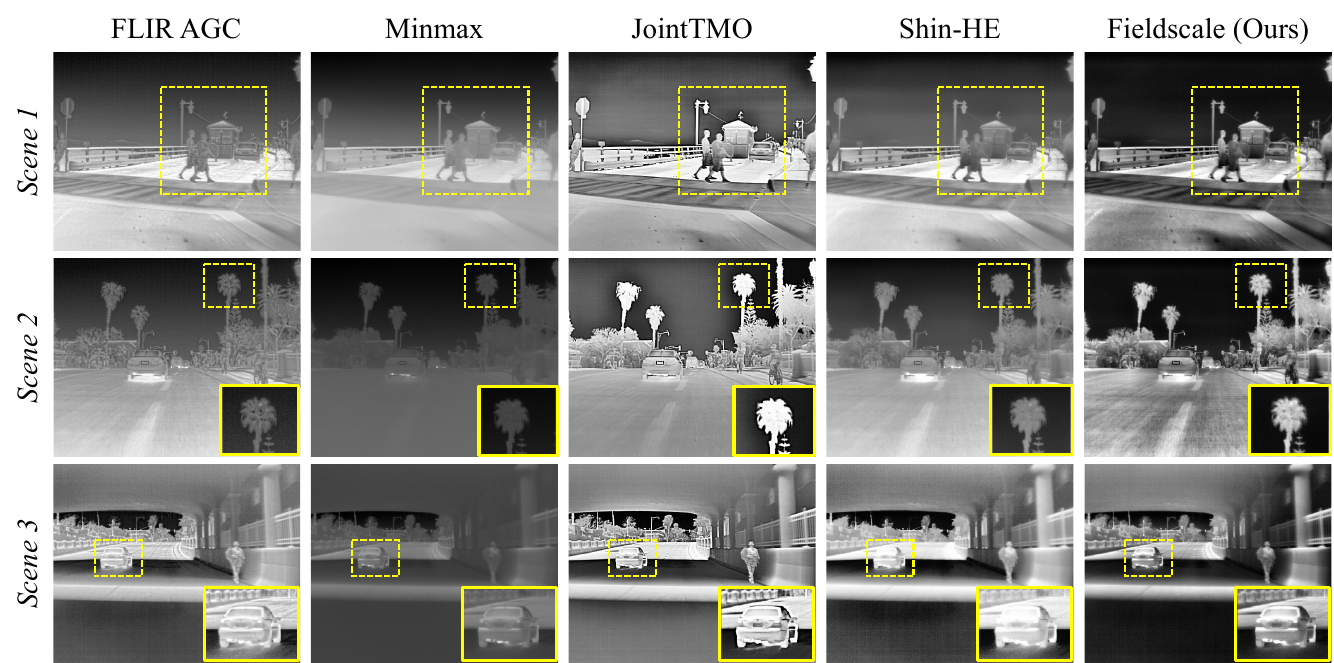}} \\
    \vspace{-2mm}
    \subfigure[\texttt{NSAVP-Day} \cite{24nsavp} + \texttt{ViViD-Night} \cite{lee2022vivid++}]{\includegraphics[width=1\linewidth]{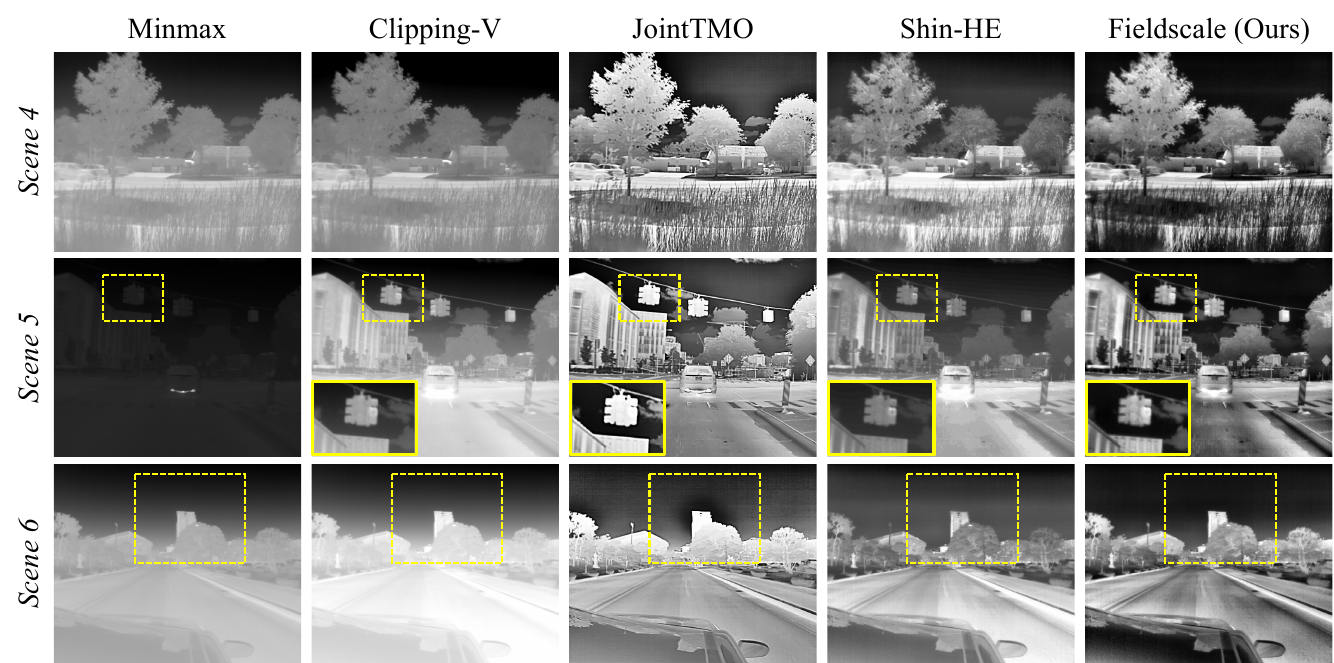}}
    \caption{Qualitative comparison of rescaled images.}
    \label{fig:iqa}
\end{figure}

\begin{figure}[!t]
    \centering
   \vspace{-1mm}
    \subfigure[\texttt{FLIR-Val} \cite{fliradas}]{\includegraphics[width=1\linewidth]{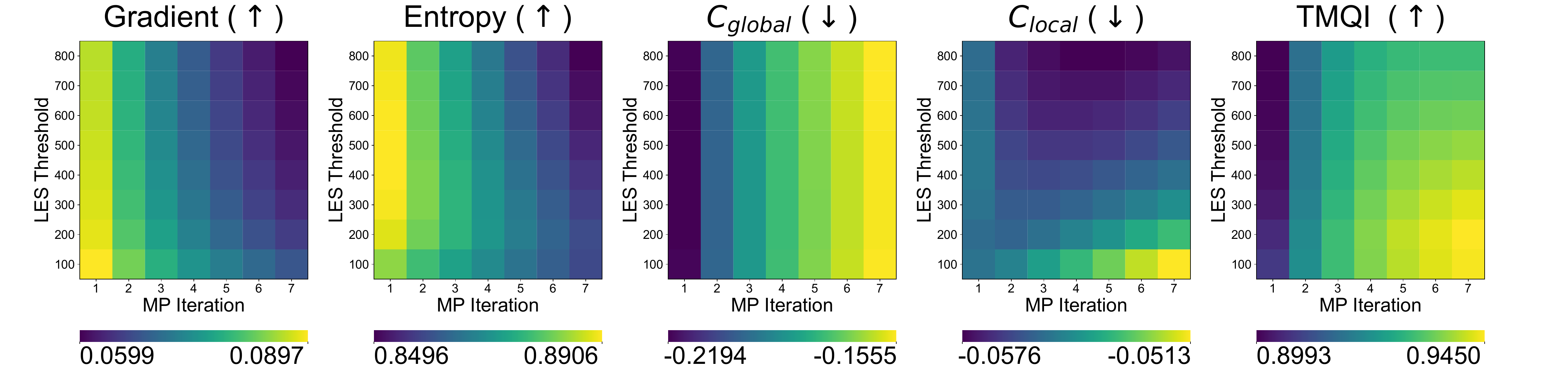}} \\
    \vspace{-1mm}
    \subfigure[\texttt{NSAVP-Day} \cite{24nsavp}]{\includegraphics[width=1\linewidth]{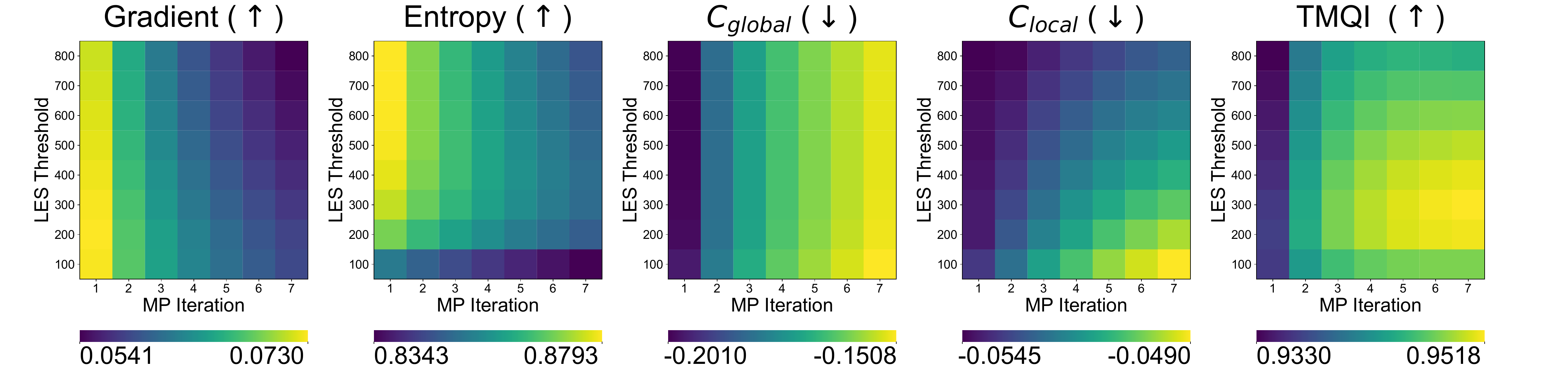}}
    \vspace{-2mm}
    \caption{IQA heatmap of different MP iterations and LES thresholds.}
    \label{fig:iqa_parameters}
    \vspace{-5mm}
\end{figure}

\tabref{tab:iqa} and \figref{fig:iqa} illustrate the quantitative and qualitative evaluation results, respectively.
FLIR AGC and Shin-HE can preserve edge information but lack sufficient contrast or suffer from the saturation of local details.
JointTMO shows boosted edge features and the dark halo at the border of foreground and background, which is typical of MSR-based approaches.
It exhibits high global contrast but relatively low naturalness, as indicated by the high $C_{\text{global}}$ and low TMQI.
Fieldscale shows the best or the second-best results on every dataset. 
It can yield images with high contrast, natural appearance, and radiometry-preserved local details in every important spatial location.

In \figref{fig:iqa_parameters}, we further analyze the influence of MP iterations $N$ and LES thresholds $T_{\text{LES}}$ on each metric. All the metrics are mainly influenced by the iteration $N$, and $T_{\text{LES}}$ regulates the sensitivity of it. For example, higher $T_{\text{LES}}$ makes the gradient more sensitive to iteration $N$.  
We observe the trade-off relationship between the loss of contrast and TMQI, where the parameters yielding the most natural images with high TMQI can decrease image contrasts.
Since the worst gradient, entropy, and loss of contrast still show better results than most of the baselines, we choose parameters that yield the best TMQI score for each downstream task. 
Training datasets of both tasks exhibit the highest score with $N=7$ and $T_{\text{LES}}=100$, so we fix the value for the remaining experiments accordingly. The ablation studies and parameter sensitivity are later analyzed in \secref{subsec:ablation}.

\subsection{Application to Night-to-Day Place Recognition}

\begin{table}[!b]
\centering
\vspace{-5mm}
\caption{Datasets used for visual place recognition}
\vspace{-1mm}
\begin{adjustbox}{width=1\linewidth}
{
\renewcommand{\arraystretch}{1.1}
\vspace{-5mm}
\begin{tabular}{clccc}
\hline
\toprule[1pt]
Split                   & Dataset   & Camera    & Image Size & \# q/db frames \\
\midrule[1pt]
\multirow{4}{*}{Train}  & ViViD++ \cite{lee2022vivid++}         & FLIR A65      & 640$\times$512    & 2753 / 5573   \\ 
                        & MS2 \cite{shin2023deep}               & FLIR A65      & 640$\times$256    & 3511 / 6836   \\ 
                        & STheReO-KAIST \cite{yun2022sthereo}   & FLIR A65      & 640$\times$512    & 1043 / 2167   \\ 
                        & Total                                 & -             & -                 & 7307 / 14576  \\
\midrule[1pt]
\multirow{4}{*}{Test}   & NSAVP-R0-F \cite{24nsavp}             & FLIR Boson    & 640$\times$512    & 1510 / 1514   \\  
                        & NSAVP-R0-R \cite{24nsavp}             & FLIR Boson    & 640$\times$512    & 1492 / 1501   \\  
                        & SNU-Winter                            & FLIR A65      & 640$\times$512    & 1198 / 1201   \\  
                        & Total                                 & -             & -                 & 4200 / 4216   \\
\bottomrule[1pt]
\end{tabular}
}
\end{adjustbox}
\label{tab:dataset}
\end{table}

\subsubsection{Setup}
Visual place recognition (VPR), a task that retrieves the same place as the query image, is chosen for our first application. Since VPR heavily depends on the overall appearance of an image, it can be an appropriate task to check if rescaled images contain sufficient global context. Considering significant temperature distribution changes between the daytime and the nighttime, we focus on night-to-day place recognition, where nighttime images are queried to retrieve the daytime images of the same scenes. To the best of our knowledge, there are no general dataset configurations for thermal place recognition (TPR) except for \cite{lee2023night}.  So, we fully exploit the public dataset and propose the TPR framework and evaluation protocols.

\subsubsection{Datasets and Baselines}
Details of the datasets are listed in \tabref{tab:dataset}. All sequences are downsampled at 5m intervals and geo-tagged with ground truth UTM coordinates acquired from RTK-GPS. Since ViViD++ \cite{lee2022vivid++}, MS2 \cite{shin2023deep}, and KAIST sequence of STheReO \cite{yun2022sthereo} span the same area in Daejeon, Korea, they are selected for training sequences and two sequences (R0-F and R0-R) of NSAVP \cite{24nsavp} for test sequences.
Noting that all public datasets are recorded in the summertime, we acquire additional winter sequences (SNU-Winter) whose trajectories are the same as the SNU sequence of STheReO and utilize them for the test. For MS2 \cite{shin2023deep}, the top and bottom 128 rows are padded with zeros to match the input size with others. While all the available daytime images are included in databases for training, only single daytime sequences of each test dataset are included for testing. Considering the real-time capability, we curate Minmax, Clipping-V \cite{23shin}, MSR \cite{97msr}, Shin-HE \cite{9804833}, and DNN-based JointTMO \cite{23joint} for baselines.

\begin{table}[!t]
\centering{
\caption{Recall Rates (Recall@1 / Recall@5) of two VPR networks (\B{bold}: the best, \ul{underline}: the second best)}
\scriptsize{
\vspace{-1mm}
\begin{tabular}{lc|c|c|c}
\toprule[1pt]
\multirow{2}{*}{Rescaling}  & 
\multicolumn{1}{c}{\multirowcell{2}{SNU\\Winter}} & 
\multicolumn{1}{c}{\multirowcell{2}{NSAVP\\R0-F \cite{24nsavp}}} & 
\multicolumn{1}{c}{\multirowcell{2}{NSAVP\\R0-R \cite{24nsavp}}} & 
\multicolumn{1}{c}{\multirowcell{2}{All\\Sequences}} \\
& \multicolumn{1}{c}{} & \multicolumn{1}{c}{} & \multicolumn{1}{c}{} & \multicolumn{1}{c}{} \\
\bottomrule[1pt]
\rowcolor{Gray}
\multicolumn{5}{c}{ResNet-18 \cite{he2016deep} + NetVLAD \cite{arandjelovic2016netvlad}} \\
\toprule[1pt]
\id{Minmax}                      & \id{88.7} / \id{96.3} & \id{93.3} / \id{97.7} & \id{81.3} / \id{89.8} & \id{85.9} / \id{93.1} \\
\id{Clipping-V \cite{23shin}}    & \ul{97.3} / \ul{99.7} & \B{97.7} / \ul{99.1} & \id{95.1} / \id{98.4} & \id{96.2} / \ul{98.9} \\
\id{Shin-HE \cite{9804833}}      & \id{97.1} / \id{99.3} & \id{93.3} / \id{97.0} & \id{96.0} / \id{98.5} & \id{93.3} / \id{97.5} \\
\id{MSR \cite{97msr}}            & \id{93.8} / \id{99.1} & \id{93.7} / \id{97.7} & \id{89.9} / \id{95.9} & \id{91.4} / \id{96.9} \\
\id{JointTMO \cite{23joint}}     & \id{96.3} / \id{99.5} & \id{96.6} / \ul{98.6} & \B{97.5} / \B{99.1} & \ul{96.5} / \ul{98.9} \\
\id{\B{Fieldscale (Ours)}}       & \B{98.3} / \B{99.8} & \ul{97.3} / \B{99.2} & \ul{97.4} / \ul{98.9} & \B{97.5} / \B{99.3} \\
\bottomrule[1pt]
\rowcolor{Gray}
\multicolumn{5}{c}{VGG16 \cite{simonyan2014very} + NetVLAD \cite{arandjelovic2016netvlad}} \\   
\toprule[1pt]
\id{Minmax}                      & \id{69.0} / \id{77.4} & \id{68.5} / \id{78.6} & \id{53.2} / \id{67.2} & \id{60.4} / \id{70.6} \\
\id{Clipping-V \cite{23shin}}    & \B{91.3} / \B{97.9} & \id{92.4} / \id{96.4} & \id{89.9} / \ul{95.9} & \id{89.8} / \id{96.0} \\
\id{Shin-HE \cite{9804833}}      & \id{91.0} / \id{97.7} & \id{82.0} / \id{92.7} & \id{81.4} / \id{92.5} & \id{81.7} / \id{92.1} \\
\id{MSR \cite{97msr}}            & \id{79.4} / \id{90.7} & \id{70.0} / \id{80.3} & \id{56.1} / \id{72.1} & \id{65.4} / \id{77.8} \\
\id{JointTMO \cite{23joint}}     & \id{86.0} / \id{95.3} & \ul{94.5} / \B{98.1} & \B{94.3} / \B{98.0} & \ul{91.1} / \B{96.7} \\
\id{\B{Fieldscale (Ours)}}       & \ul{91.1} / \ul{97.8} & \B{95.1} / \ul{97.5} & \ul{90.2} / \ul{95.9} & \B{91.3} / \ul{96.6} \\
\bottomrule[1pt]
\vspace{-5mm}
\end{tabular}
\label{tab:recall_summer}
}
}
\end{table}

\begin{figure}[!t]
    \centering
   \includegraphics[width=0.9\columnwidth]{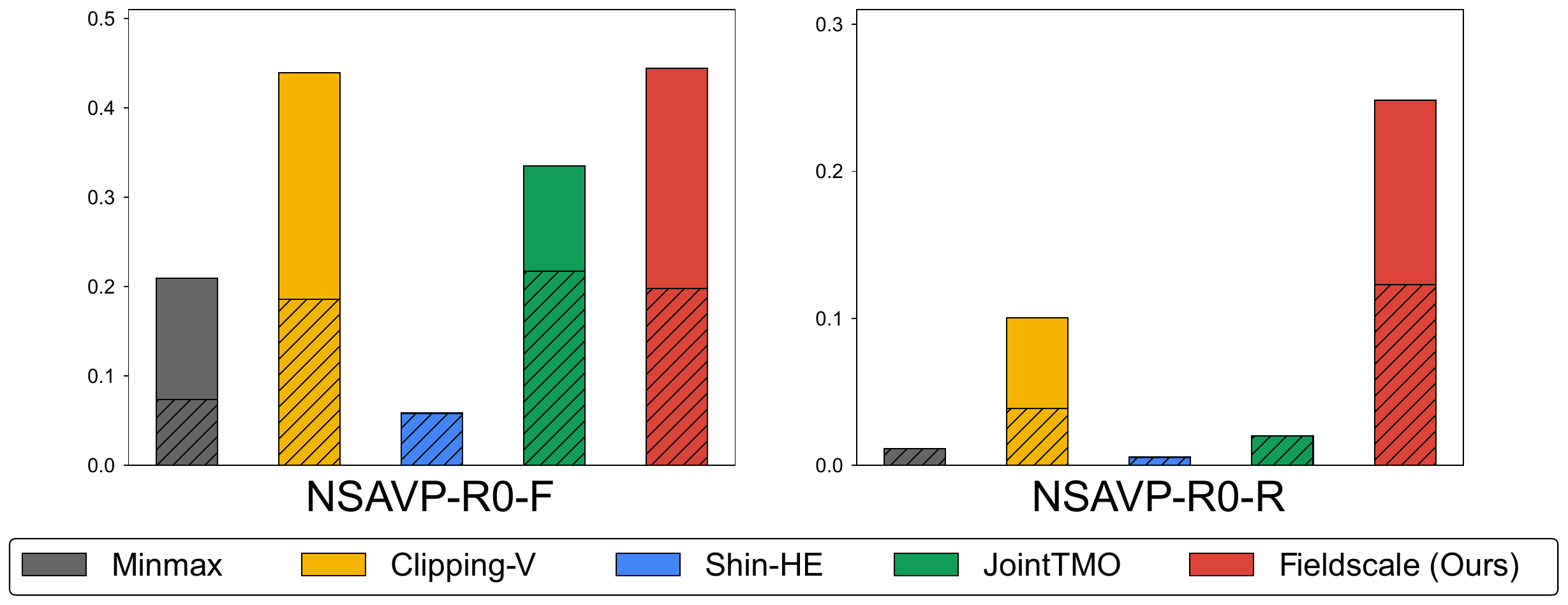}
    \vspace{-1mm}
    \caption{
    Recall at 100\% precision (marked with diagonal lines) and 99\% precision comparison within VGG16 \cite{simonyan2014very} + NetVLAD \cite{arandjelovic2016netvlad}. 
    While baseline methods with high recall rates exhibit low precisions, 
    Fieldscale shows high precision for both sequences.
    }
    \label{fig:recallat99}
    \vspace{-5mm}
\end{figure}

\subsubsection{Implementation Details}
Our place recognition network consists of NetVLAD \cite{arandjelovic2016netvlad} with 64 clusters, one of the most commonly used feature aggregation modules, following two different backbones (ResNet-18 \cite{he2016deep} and VGG16 \cite{simonyan2014very}) with ImageNet pre-trained weights. Triplet mining strategies and most of the hyperparameters are adapted from the default settings of the existing benchmarks \cite{berton2022deep}. Adam (1e-5 learning rate and no weight decay) is selected for the optimizer, and no scheduler is utilized. No data augmentation is included, and the training batch size is set to 4. Training is stopped early when recall@5 is not improved more than five times after 20 epochs. Following the standard protocols of VPR literature \cite{berton2022deep}, recall@N is selected as the primary metric, and up to 25m difference of geo-tagged poses are regarded as correct recalls. Subsequent gamma correction and CLAHE are removed since maximizing the local contrast is unnecessary for global appearance-dependent VPR.

\subsubsection{Discussion}
Recall rates are summarized in \tabref{tab:recall_summer}. 
ResNet backbone shows better results than VGG16, which aligns with the benchmark results \cite{berton2022deep}. 
Among competitive baselines, Fieldscale consistently exhibits the best or second-best recall rates for all sequences.
JointTMO struggles on SNU-Winter but exhibits promising results on the NSAVP R0-R sequence, especially with the VGG16 backbone, where the gap becomes minor with the ResNet backbone.
While recall rates are essential metrics from an image retrieval perspective, retrieval precision is crucial to loop closure detection (LCD) in \ac{SLAM}. 
\figref{fig:recallat99} plots recall at 100\% and 99\% precision, widely chosen metrics to verify the precision of VPR \cite{tipaldi2013geometrical, zaffar2021vpr}. 
Baselines with competing recall@1 results show much lower precision than Fieldscale, especially in the R0-R sequence.
It can be concluded that Fieldscale can help networks learn better representation by distinguishing different places with similar appearances and imposing higher similarities on the same places.

\subsection{Application to Object Detection}
Object detection, the problem of locating semantic objects in given images, is chosen for our second application to evaluate whether rescaled images preserve local details.

\subsubsection{Datasets and Baselines}
Following the previous work \cite{munir2021sstn}, experiments are conducted on the FLIR ADAS dataset \cite{fliradas} with three classes (pedestrian, car, and bicycle). Minmax, MSR \cite{97msr}, JointTMO \cite{23joint}, Shin-HE \cite{9804833}, and FLIR AGC are chosen for the baseline. Four object detection networks (YOLOX-M \cite{ge2021yolox}, ATSS \cite{zhang2020bridging}, sparse R-CNN \cite{sun2021sparse} and cascade R-CNN \cite{cai2018cascade}) are utilized and their mean average precision (mAP) with COCO \cite{lin2014microsoft} format are reported.

\begin{table}[!t]
\centering
\caption{Object detection results from different models}
\scriptsize{
\vspace{-1mm}
\begin{adjustbox}{width=1\linewidth}
\renewcommand{\arraystretch}{1} 
\begin{tabular}{lcccccc}
\toprule[1pt]
Rescaling            & \text{AP}    & $\text{AP}_{50}$  & $\text{AP}_{75}$  & $\text{AP}_{s}$   & $\text{AP}_{m}$   & $\text{AP}_{l}$\\
\bottomrule[1pt]
\rowcolor{Gray}
\multicolumn{7}{c}{YOLOX-M \cite{ge2021yolox}} \\
\toprule[1pt]
FLIR AGC                & \id{48.4}    & \id{83.5}    & \id{47.8}    & \id{35.1}    & \id{55.6}    & \id{66.1} \\ 
Minmax                  & \id{47.0}    & \id{82.0}    & \id{45.5}    & \id{33.2}    & \id{54.4}    & \id{67.6} \\ 
MSR \cite{97msr}        & \id{47.9}    & \id{83.5}    & \id{46.3}    & \id{34.2}    & \id{55.1}    & \id{66.1} \\ 
JointTMO \cite{23joint} & \id{48.3}    & \id{83.7}    & \id{47.9}    & \id{35.5}    & \id{55.0}    & \id{65.9} \\ 
Shin-HE \cite{9804833}  & \id{47.8}    & \id{83.3}    & \id{47.1}    & \id{34.2}    & \id{54.9}    & \id{65.6} \\ 
\B{Fieldscale (Ours)}   & \BK{49.1}    & \BK{84.4}    & \BK{48.4}    & \BK{36.1}    & \BK{55.8}    & \BK{68.1} \\ 
\bottomrule[1pt]
\rowcolor{Gray}
\multicolumn{7}{c}{ATSS \cite{zhang2020bridging}} \\
\toprule[1pt]
FLIR AGC                & \id{44.6}    & \id{82.1}    & \id{42.4}    & \id{34.2}    & \id{51.0}    & \id{59.3} \\
Minmax                  & \id{44.6}    & \id{81.1}    & \id{42.2}    & \id{32.5}    & \id{52.3}    & \id{59.3} \\
MSR \cite{97msr}        & \id{44.9}    & \id{81.3}    & \id{42.0}    & \id{33.2}    & \id{52.0}    & \id{61.5} \\
JointTMO \cite{23joint} & \id{45.0}    & \id{82.1}    & \id{41.9}    & \id{34.2}    & \id{51.7}    & \id{54.6} \\
Shin-HE \cite{9804833}  & \id{45.2}    & \id{82.0}    & \id{43.4}    & \id{33.4}    & \id{51.9}    & \id{61.2} \\
\B{Fieldscale (Ours)}   & \BK{46.6}    & \BK{83.0}    & \BK{45.4}    & \BK{34.8}    & \BK{53.5}    & \BK{63.8} \\
\bottomrule[1pt]
\rowcolor{Gray}
\multicolumn{7}{c}{Sparse R-CNN \cite{sun2021sparse}} \\
\toprule[1pt]
FLIR AGC                & \id{41.4}    & \id{77.9}   & \id{38.2}    & \id{31.4}    & \id{46.9}    & \id{58.0} \\
Minmax                  & \id{40.6}    & \id{76.7}   & \id{37.4}    & \id{30.4}    & \id{46.7}    & \id{55.9} \\
MSR \cite{97msr}        & \id{39.2}    & \id{72.8}   & \id{36.3}    & \id{28.6}    & \id{45.6}    & \id{52.8} \\
JointTMO \cite{23joint} & \id{41.7}    & \id{79.0}   & \id{38.2}    & \id{31.7}    & \id{47.9}    & \id{56.2} \\
Shin-HE \cite{9804833}  & \id{41.1}    & \id{77.7}   & \id{37.5}    & \id{31.4}    & \id{47.0}    & \id{57.1} \\
\B{Fieldscale (Ours)}   & \BK{42.9}    & \BK{79.2}   & \BK{40.7}    & \BK{32.0}    & \BK{49.3}    & \BK{60.6} \\
\bottomrule[1pt]
\rowcolor{Gray}
\multicolumn{7}{c}{Cascade R-CNN \cite{cai2018cascade}} \\
\toprule[1pt]
FLIR AGC                & \id{44.1}    & \id{80.0}    & \id{43.1}    & \id{32.7}    & \id{51.5}    & \id{58.7} \\
Minmax                  & \id{43.6}    & \id{78.7}    & \id{41.3}    & \id{31.4}    & \id{50.7}    & \id{58.6} \\
MSR \cite{97msr}        & \id{44.7}    & \id{79.3}    & \id{43.5}    & \id{33.2}    & \id{51.3}    & \id{61.6} \\
JointTMO \cite{23joint} & \id{45.2}    & \id{82.2}    & \id{43.1}    & \id{33.6}    & \id{51.4}    & \id{62.2} \\
Shin-HE \cite{9804833}  & \id{43.9}    & \id{80.8}    & \id{41.2}    & \id{31.7}    & \id{50.4}    & \id{61.6} \\
\B{Fieldscale (Ours)}   & \BK{46.5}    & \BK{82.2}    & \BK{45.3}    & \BK{35.2}    & \BK{52.7}    & \BK{63.7} \\
\bottomrule[1pt]
\end{tabular}
\end{adjustbox}
\label{tab:detection_all}
}
\end{table}

\subsubsection{Implementation Details}
All models are trained for 1x schedule (12 epochs) with pre-trained weights available at the codebase\footnote{https://github.com/open-mmlab/mmdetection}. Random horizontal flipping and resizing are used for data augmentation. 
For YOLOX-M, an initial learning rate of 0.01 is scheduled with the cosine lr scheduler, and stochastic gradient descent (SGD) with 0.9 momentum and 0.0005 weight decay is used for optimizer. 
The rest of the models use ResNet-50 \cite{he2016deep} for the backbone.
For the optimizer, sparse R-CNN uses AdamW with a 2.5e-5 initial learning rate and 1e-4 weight decay, and both ATSS and Cascade R-CNN use SGD with a 0.02 initial learning rate, 0.9 momentum, and 1e-4 weight decay.
The learning rate is scheduled to be decayed at epochs 8 and 11 by a factor of 0.1. 
Every training is done on a single RTX 3090 GPU with a batch size of 8 or 16, depending on the model size.

\subsubsection{Discussion}
As reported in \tabref{tab:detection_all}, Fieldscale shows the highest AP for every IoU threshold and box size compared to all other baselines, including FLIR AGC. Notably, it significantly improves $\text{AP}_{s}$, demonstrating that contrasts and details in a small area can be well preserved.

\begin{figure}[!t]
\centering
\includegraphics[width=0.9\columnwidth]{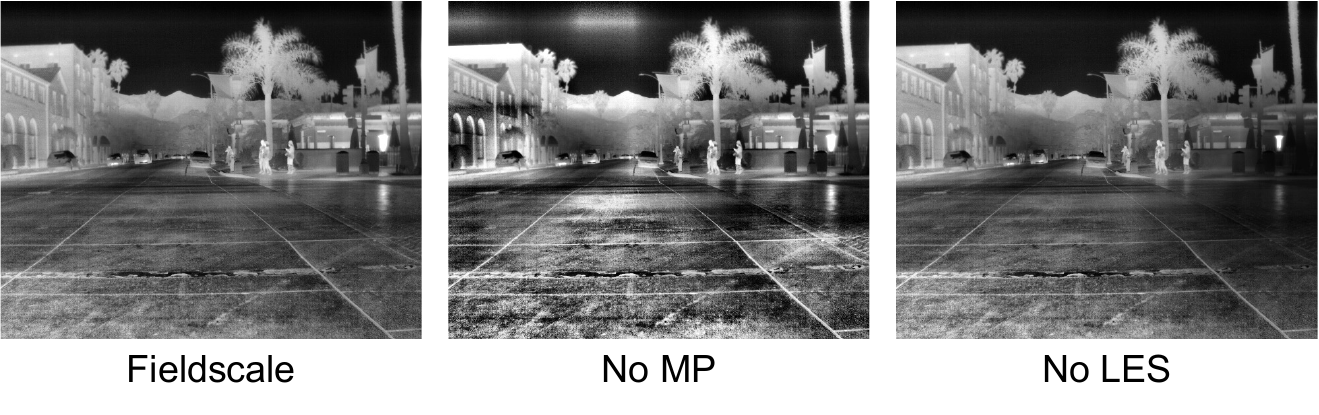}
\vspace{-1mm}
\caption{
Qualitative example of ablating each module of Fieldscale. 
MP removes the tiling artifacts and smooths the image, and LES hinders hot objects from darkening the local area of an image.
Both modules should exist to generate the best result.
}
\label{fig:ablation_iqa}
\vspace{-3mm}
\end{figure}

\begin{table}[!t]
\centering
\caption{Ablation studies on LES and MP modules}
\begin{adjustbox}{width=1\linewidth}
\renewcommand{\arraystretch}{1.2} 
\begin{tabular}{@{\extracolsep{5pt}}ccccc}
\toprule
\multirow{2}{*}{LES}    & \multirow{2}{*}{MP}   & \multicolumn{1}{c}{IQA} & \multicolumn{1}{c}{Object Detection}  & \multicolumn{1}{c}{VPR} \\
                        \cline{3-3} \cline{4-4} \cline{5-5}
                        &                        & $C_{\text{global}}$ / $C_{\text{local}}$ / TMQI & $\text{AP}$ / $\text{AP}_{50}$ / $\text{AP}_{75}$ & R@1 / R@5\\
\midrule
           &            & -0.2377 / -0.0518 / 0.8969    & 45.5 / 79.9 / {44.5}    & 94.5 / 98.0 \\
\checkmark &            & -0.2333 / -0.0534 / 0.9129    & 45.8 / 80.5 / {44.6}    & 96.3 / 98.9 \\
           & \checkmark & -0.1971 / -0.0552 / 0.9175    & 46.2 / 81.7 / {45.3}    & 95.5 / 98.2 \\
\checkmark & \checkmark & -0.1508 / -0.0490 / 0.9481    & 46.5 / 82.2 / {45.3}    & 97.5 / 99.3 \\

\bottomrule
\vspace{-5mm}
\end{tabular}%
\end{adjustbox}
\label{tab:ablation_modules}
\end{table}

\begin{table}[!t]
\centering
\caption{Parameters sensitivity analysis }
\vspace{-1mm}
\begin{adjustbox}{width=1\linewidth}
\renewcommand{\arraystretch}{1.1} 
    \subfigure[MP Iteration ($N$)]{
        \begin{tabular}{@{\extracolsep{3pt}}ccc}
            \toprule[1pt]
            \multirow{2}{*}{$N$} & Object Detection & VPR \\ 
            \cline{2-2} \cline{3-3}
            & $\text{AP}$ / $\text{AP}_{50}$ / $\text{AP}_{75}$ & R@1 / R@5  \\ 
            \midrule
            1   & 46.4 / 82.2 / 45.0  & 97.0 / 99.1 \\
            3   & 45.5 / 81.9 / 45.3  & 97.2 / 99.1 \\
            5   & 46.5 / 81.9 / 45.5  & 97.5 / 99.3 \\
            7   & 46.5 / 82.2 / 45.3  & 97.5 / 99.3 \\
            \bottomrule[1pt]
            \vspace{-5mm}
        \end{tabular}   
        }
    \subfigure[LES Threshold ($T_{\text{LES}}$)]{
        \begin{tabular}{@{\extracolsep{3pt}}ccc}
            \toprule[1pt]
            \multirow{2}{*}{$T_{\text{LES}}$} & Object Detection & VPR \\ 
            \cline{2-2} \cline{3-3}
            & $\text{AP}$ / $\text{AP}_{50}$ / $\text{AP}_{75}$ & R@1 / R@5  \\ 
            \midrule
            100   & 46.5 / 82.2 / 45.3 & 97.5 / 99.3 \\
            200   & 46.7 / 82.3 / 45.6 & 97.5 / 99.3 \\
            400   & 46.6 / 82.1 / 45.2 & 97.4 / 99.3 \\
            800   & 46.6 / 81.9 / 45.3 & 97.5 / 99.2 \\
            \bottomrule[1pt]
            \vspace{-5mm}
        \end{tabular}    
        }
\end{adjustbox}
\vspace{-5mm}
\label{tab:ablation_parameters}
\end{table}

\subsection{Ablation Study}
\label{subsec:ablation}
\figref{fig:ablation_iqa} illustrates the effect of ablating two core modules, LES and MP.
LES can suppress the influence of local hot objects on global appearance, and MP can yield smoother images.
\tabref{tab:ablation_modules} and \tabref{tab:ablation_parameters} provide quantitative analysis of our ablation studies and parameter sensitivities, 
where \texttt{NSAVP-Day} \cite{24nsavp} is used for IQA analysis, Cascade R-CNN \cite{cai2018cascade} for object detection, and ResNet-NetVLAD \cite{he2016deep, arandjelovic2016netvlad} for VPR.
While LES and MP may weaken the contrast, they can generate more natural images, as shown in the increased TMQI. 
Using LES or MP alone yields suboptimal results, but combining both modules can significantly boost performance.
Once both of them are provided, performance changes are minor, demonstrating parameter insensitivity and guaranteeing that parameter values can be roughly chosen without specialized domain knowledge.

\subsection{Time Analysis}
\begin{figure}[!t]
    \centering
    \subfigure[Grid Size]{
		\includegraphics[width=0.45\columnwidth]{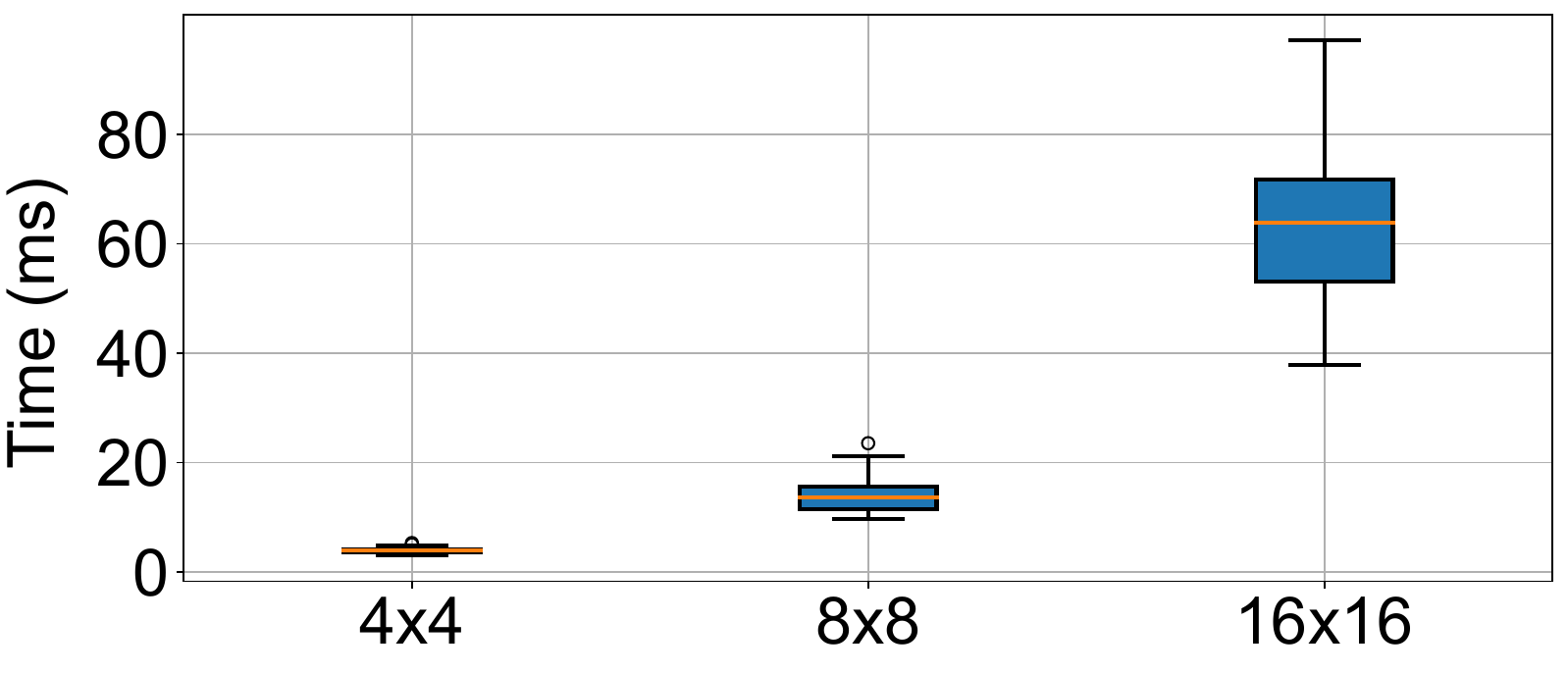}}
    \subfigure[MP Iteration ($N$)]{
		\includegraphics[width=0.45\columnwidth]{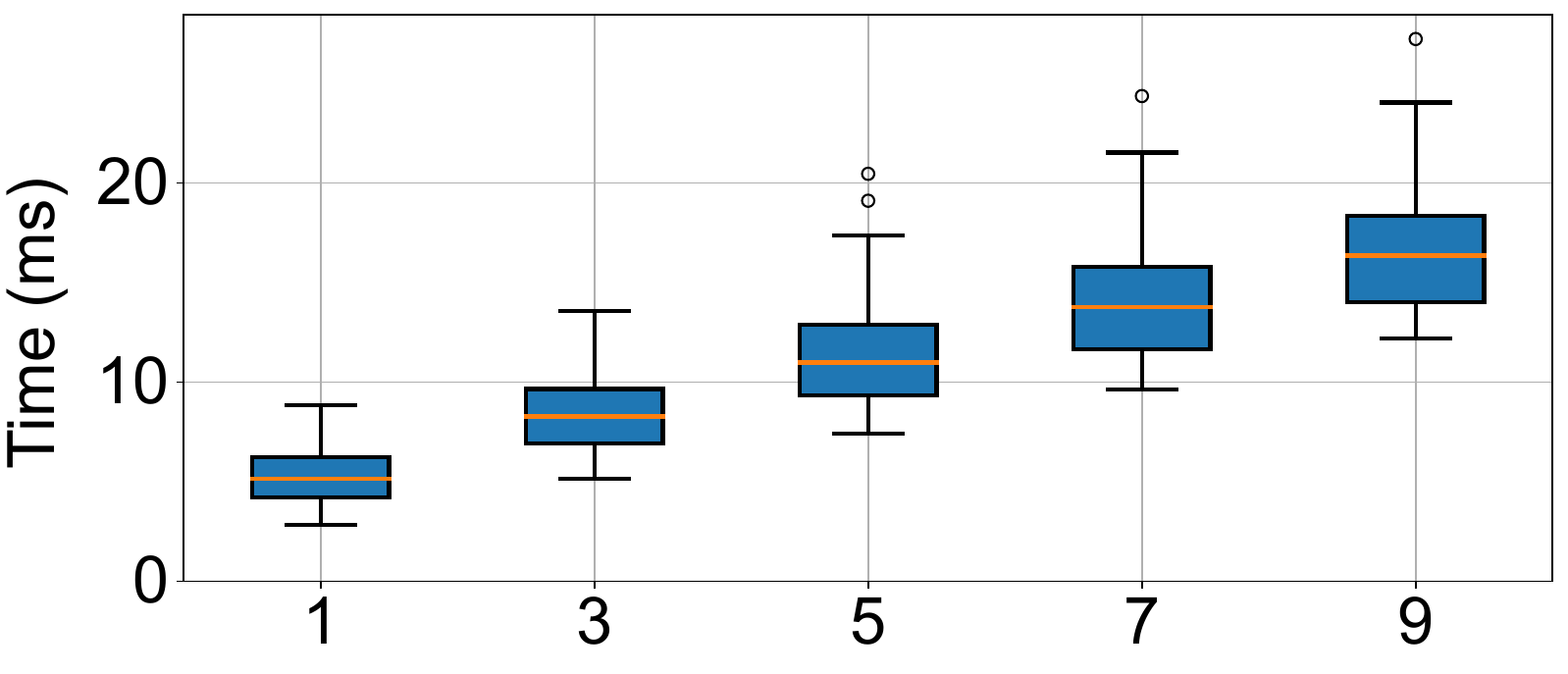}} \\
    \subfigure[Local Distance ($d$)]{
		\includegraphics[width=0.45\columnwidth]{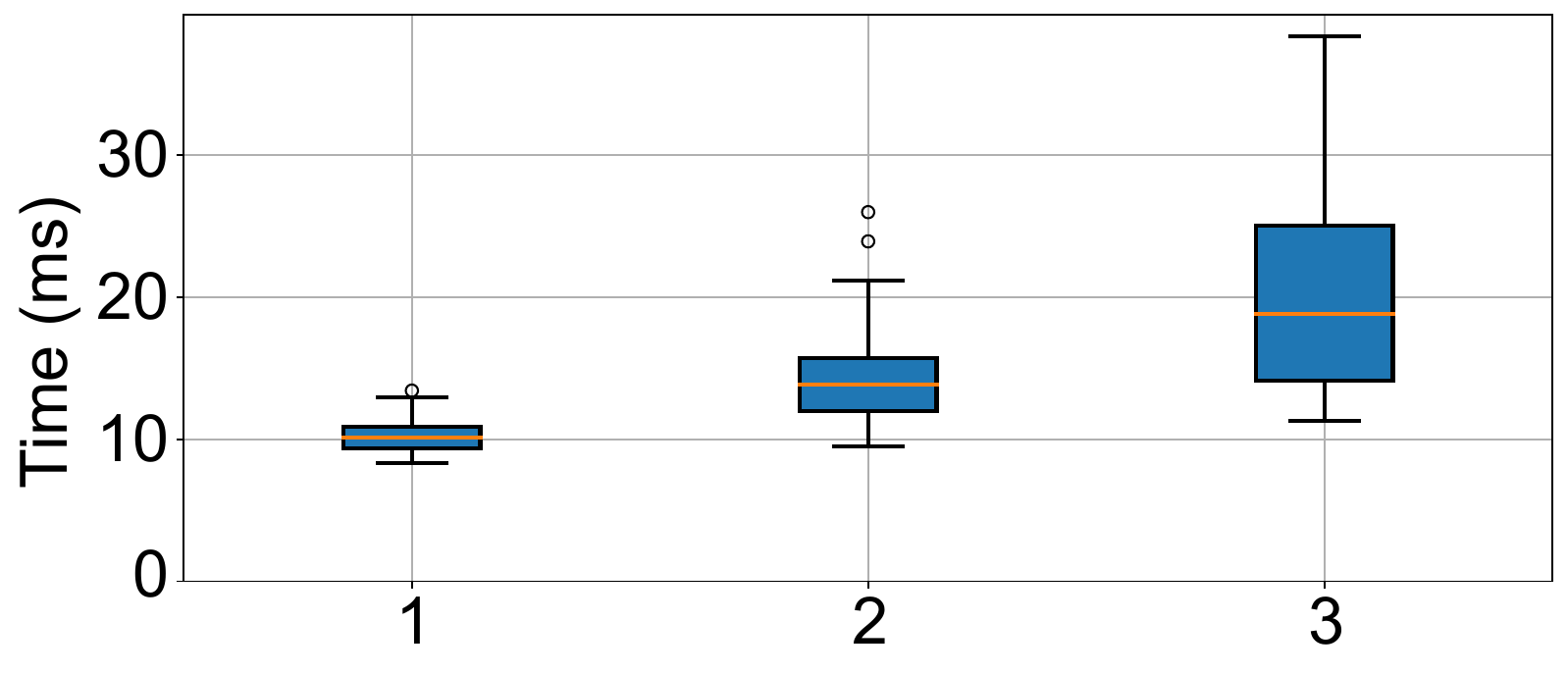}}
    \subfigure[LES Threshold ($T_{\text{LES}}$)]{
		\includegraphics[width=0.45\columnwidth]{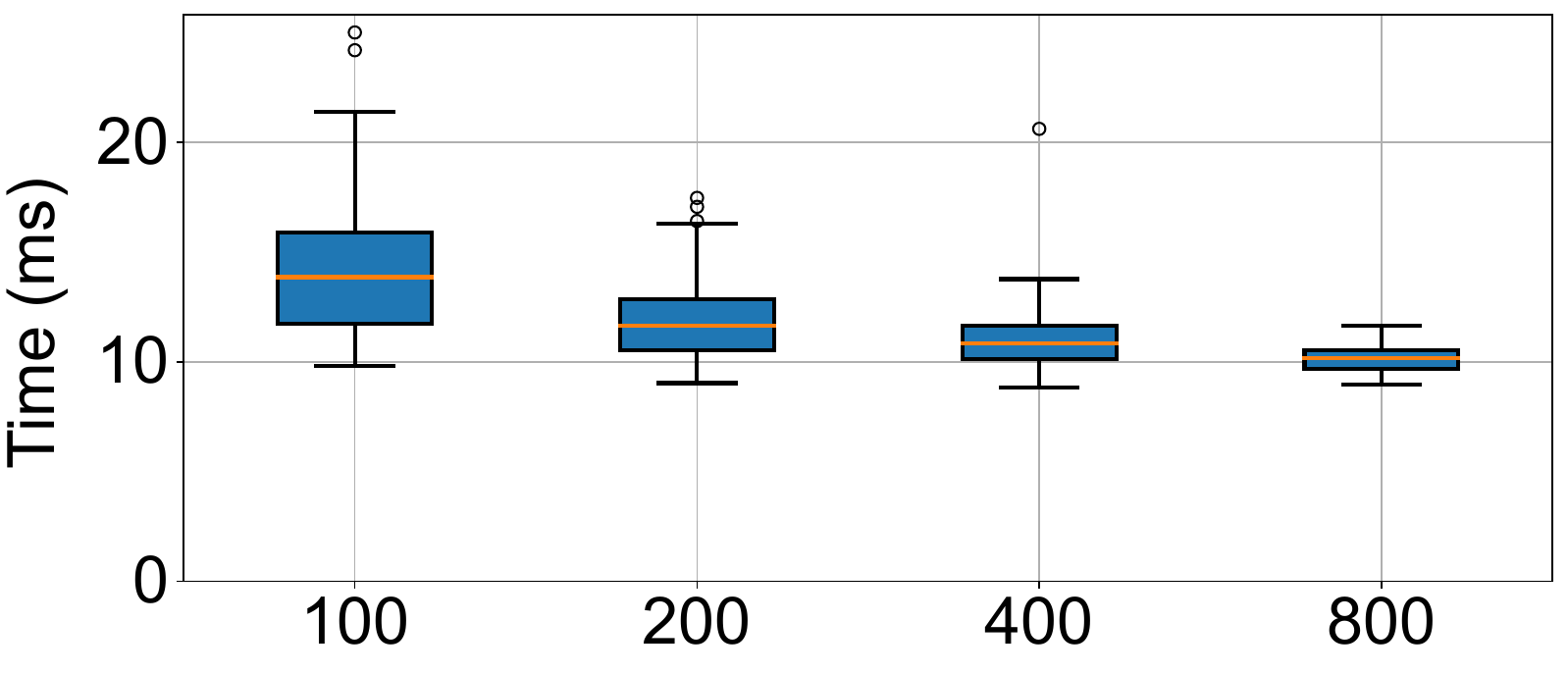}} \\
    \vspace{-1mm}
    \caption{Influence of each parameter on field construction time.}
    \label{fig:time_analysis}
    \vspace{-5mm}
\end{figure}

As real-time capability is one of the most important aspects of robotics applications,  we check the time it takes for our system to rescale RAW images in \tabref{tab:time_analysis}. 10\% of images are subsampled from a mixture of \texttt{FLIR-Val} and \texttt{NSAVP-Day}. Considering that our default setting is the slowest option, both DEFAULT setting ($N$=7, $T_{\text{LES}}$=100) and FAST setting ($N$=1, $T_{\text{LES}}$=800) are considered. The FAST setting is still viable since \tabref{tab:ablation_parameters} shows similar performance regardless of parameter values. Field construction can be executed in real-time (72.3Hz and 323.6Hz) with Intel i7-12700 for both settings, and can be improved with additional parallel computing. Furthermore, field construction times after subtle parameter changes are plotted on \figref{fig:time_analysis}, in which any combination of parameters will take less than the average 20ms (50Hz). The only exception is the $16\times16$ grid size, which is not recommended because each patch becomes too small and produces sub-optimal performance.

\begin{table}[!t]
\centering
\caption{Computation time analysis (\textit{mean $\pm$ std. dev.}) }
\vspace{-1mm}
\begin{adjustbox}{width=1\linewidth}
\renewcommand{\arraystretch}{1.1} 
\begin{tabular}{cccc}
\toprule[1pt]
Setting &   Field Construction & Field-based Rescaling & Total                    \\
\midrule[1pt]
DEFAULT     & 13.83 $\pm$ 2.64 ms    & 5.60 $\pm$ 0.70 ms       & \textbf{19.43 $\pm$ 2.70 ms}\\
FAST        & 3.09 $\pm$ 0.61 ms     & 5.45 $\pm$ 0.95 ms      & \textbf{8.53 $\pm$ 1.29 ms}\\
\bottomrule[1pt]
\end{tabular}
\end{adjustbox}
\label{tab:time_analysis}
\vspace{-2mm}
\end{table}

\subsection{Limitations and Future Works}
Although Fieldscale can provide strong \textit{spatial consistency} within an image, \textit{temporal consistency} may be required for video-related tasks \cite{saputra2021graph, chen2023thermal}.
Fieldscale can yield consistent results if each object's local neighborhood remains the same, but local flickering or tiling artifacts can occur when hot objects in one grid move to adjacent grids or outside the frame.
We note that local flickering, a common aspect of tile-based image processing methods \cite{15realtime}, has little influence on global contrast and thus hardly affects the performance of single image-based tasks.
One simple yet effective solution is to smooth current fields with the past, as fields change slower than the image pixel.
However, more sophisticated solutions can be implemented, and we leave this to our future work.

\section{Conclusion}
\label{sec:conclusion}

Fieldscale addresses RAW TIR image rescaling, the primary and essential step for handling TIR cameras in robotics. It effectively captures thermal locality, yielding rescaled images with maximized information and naturalness. The potential of Fieldscale lies in its real-time processing capability, scalability of each component, and resilience to varying parameters. It can boost performance in a diverse range of TIR image-based applications. In the future, we are planning to improve the temporal consistency of Fieldscale within thermal videos, ultimately suggesting a rescaling framework with both spatial and temporal consistency.

\balance
\footnotesize
\bibliographystyle{IEEEtranN}
\bibliography{string-short, references}

\end{document}